\documentclass{article}

\PassOptionsToPackage{numbers, sort}{natbib}
\usepackage[preprint]{neurips_2023}

\usepackage[utf8]{inputenc} %
\usepackage[T1]{fontenc}    %
\usepackage{hyperref}       %
\usepackage{url}            %
\usepackage{booktabs}       %
\usepackage{amsfonts}       %
\usepackage{nicefrac}       %
\usepackage{microtype}      %
\usepackage{xcolor}         %
\usepackage{graphicx}
\usepackage{enumitem}
\usepackage{multirow}
\usepackage{wrapfig}
\usepackage{booktabs}
\usepackage{caption}
\usepackage{subcaption}
\usepackage{algorithm}
\usepackage{algpseudocode}

\usepackage{amsmath}
\usepackage{amssymb}
\usepackage{mathtools}
\usepackage{amsthm}

\title{Detector Guidance for Multi-Object \\ Text-to-Image Generation}

\author{%
  Luping Liu$^1$, Zijian Zhang$^1$, Yi Ren$^2$, Rongjie Huang$^1$, Xiang Yin$^2$, Zhou Zhao$^1$\thanks{Corresponding author}\\
  $^1$Zhejiang University\ \  $^2$ByteDance AI Lab\\
  \texttt{\{luping.liu, ckczzj, rongjiehuang, zhaozhou\}@zju.edu.cn,} \\\texttt{\{ren.yi, yinxiang.stephen\}@bytedance.com}
   \\
}

\begin{document}

\maketitle

\begin{abstract}
  Diffusion models have demonstrated impressive performance in text-to-image generation. They utilize a text encoder and cross-attention blocks to infuse textual information into images at a pixel level. However, their capability to generate images with text containing multiple objects is still restricted. Previous works identify the problem of information mixing in the CLIP text encoder and introduce the T5 text encoder or incorporate strong prior knowledge to assist with the alignment. We find that mixing problems also occur on the image side and in the cross-attention blocks. The noisy images can cause different objects to appear similar, and the cross-attention blocks inject information at a pixel level, leading to leakage of global object understanding and resulting in object mixing. In this paper, we introduce Detector Guidance (DG), which integrates a latent object detection model to separate different objects during the generation process. DG first performs latent object detection on cross-attention maps (CAMs) to obtain object information. Based on this information, DG then masks conflicting prompts and enhances related prompts by manipulating the following CAMs. We evaluate the effectiveness of DG using Stable Diffusion on COCO, CC, and a novel multi-related object benchmark, MRO. Human evaluations demonstrate that DG provides an 8-22\% advantage in preventing the amalgamation of conflicting concepts and ensuring that each object possesses its unique region without any human involvement and additional iterations. Our implementation is available at \url{https://github.com/luping-liu/Detector-Guidance}.
  
\end{abstract}

\section{Introduction}

Diffusion models \citep{sohl2015deep, ho2020denoising,song2021denoising,song2021scorebased} have exhibited impressive performance in conditional generations, which requires that the generation results be not only realistic but also strongly correlated with the given conditions. Among various conditions, text condition has attracted significant attention due to its user-friendly nature and has resulted in a plethora of heavyweight works, such as DALL$\cdot$E 2 \citep{ramesh2022hierarchical}, Imagen \citep{saharia2022photorealistic} and Stable Diffusion \citep{rombach2022high}. By utilizing billions of text-image pairs from the internet and employing well-designed model structures, these models ultimately achieve state-of-the-art text-to-image performance.

However, these models still exhibit relatively poor performance in generating multiple objects within a single image. Issues such as attribute mixing, object mixing, and object disappearance persist. Attribute mixing refers to the phenomenon where objects may be influenced by attributes that belong to other objects. Object mixing and disappearance refer to the fusion that occurs at the object level, leading to the generation of strange multi-object hybrids and incorrect object count.

\begin{figure}[t]
  \centering
  \includegraphics[width=\linewidth, trim=0 15 0 0]{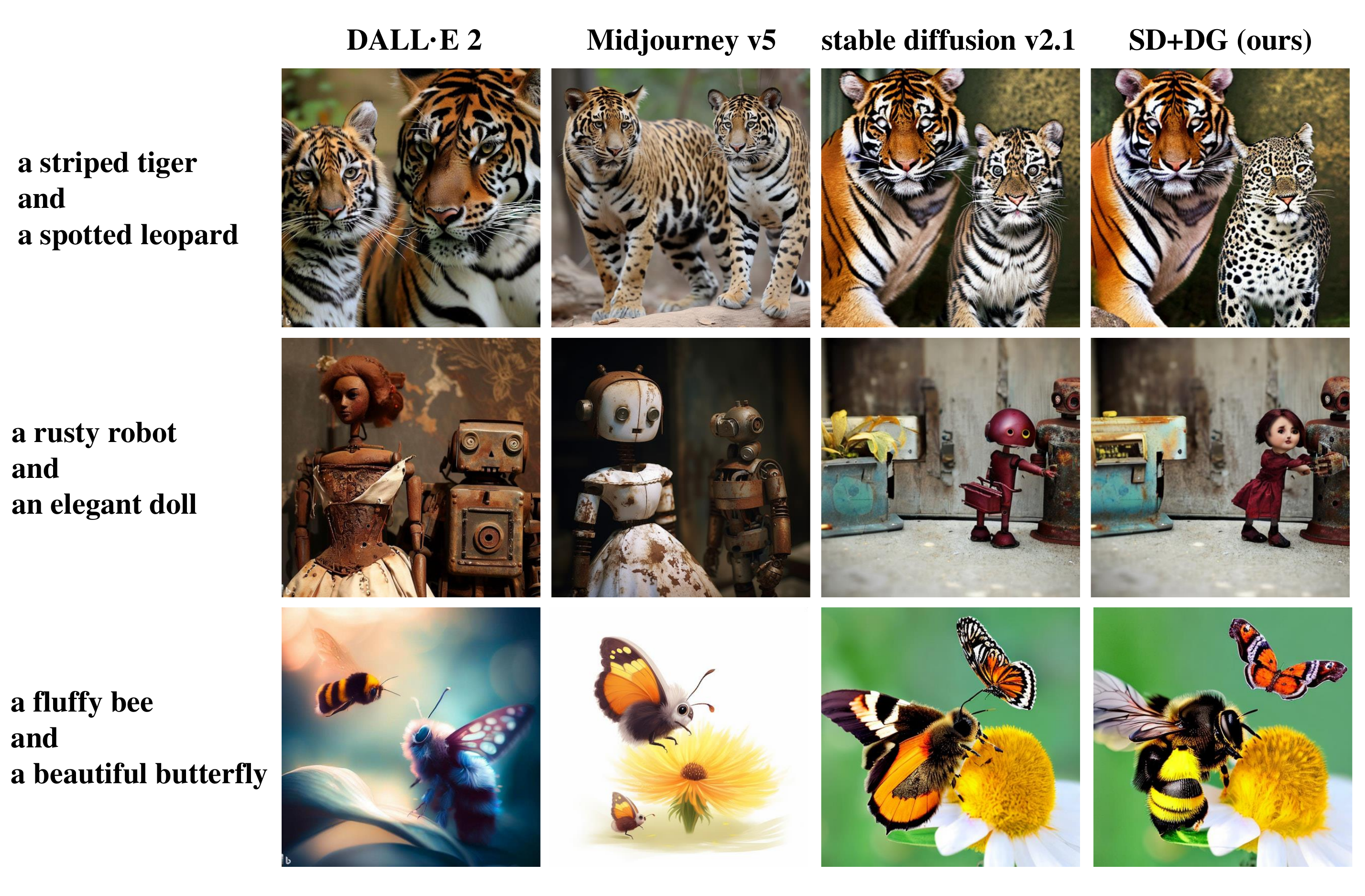}
  \caption{Examples of the object mixing problem in text-to-image diffusion models can be observed in DALL$\cdot$E, Midjourney v5, and Stable Diffusion 2.1. This issue can be resolved in Stable Diffusion 2.1 through the implementation of our Detector Guidance.}
  \label{fig_compare1}
  \vspace*{-0.5cm}
\end{figure}

Prior research \citep{feng2022training} has revealed that due to the causal attention masks in the text encoder, the semantics of tokens in the later part of a sequence get mixed with the token semantics before them. We further discover that a similar information mixing issue arises on the image side. The intermediate results of the diffusion model contain noise, which can cause different objects to appear similar. These two problems increase the difficulty of aligning different objects in prompts to the correct regions. Furthermore, diffusion models employ cross-attention blocks between text and images to incorporate text conditions into each pixel. In situations where the text conditions contain multiple objects, this creates a pixel-by-pixel competition for information from different objects. This can result in the fusion of conflicting information, such as 40\% leopard and 60\% tiger (e.g., the 1st row in Figure \ref{fig_compare1}), or the division of a complete region by texts from different objects (e.g., the 3rd row in Figure \ref{fig_compare1}). This underscores the weak global comprehension abilities of cross-attention blocks. %

Previous works solve this problem by incorporating strong prior knowledge, improving the correspondence between attributes and objects, or introducing better text encoders. The prior knowledge may include bounding boxes \citep{ma2023directed}, masks \citep{avrahami2022spatext}, or small patches \citep{sarukkai2023collage} of target objects. These data can aid cross-attention in achieving better alignment between the target prompt and image patches, thereby reducing undesired mixing. However, such solutions necessitate extensive human intervention and restrict the diversity of generation results. \citet{feng2022training} utilize language parsers to associate attributes solely with the corresponding objects. However, this method is effective only when there is no issue of object mixing. \citet{saharia2022photorealistic} utilize the T5 text encoder instead of CLIP, but it cannot address the problems on the image side and in the cross-attention blocks. %

In this paper, our solution is to enable diffusion models to grasp the concept of objects, allowing them to assign regions globally and generate different objects simultaneously. To achieve this objective, we integrate a latent object detection model into pre-trained diffusion models. During the generation process, the latent object detection model generates bounding boxes based on the cross-attention maps (CAMs). By selecting CAMs as input, we can fully utilize the alignment results of the diffusion model, which increases the robustness and generalization of the detection. Once we obtain object information, we combine the bounding boxes and CAMs to further refine the boundary and build masks. Then we mask conflicting text prompts and enhance the target text prompts by manipulating CAMs. Additionally, we use a smooth strategy to ensure continuity and support high-order numerical methods. We refer to our approach as detector guidance (DG). 

We evaluate the effectiveness of DG using Stable Diffusion on COCO \citep{lin2014microsoft}, CC \citep{feng2022training}, and a new multi-related object benchmark (MRO). Based on our experiments, DG outperforms the original Stable Diffusion by 8-22\% in human evaluation. DG accurately assigns attributes to the corresponding objects, prevents the combination of conflicting concepts, and ensures that each object has its unique region due to its clear understanding of objects. Our paper has the following contributions:

\begin{itemize}[leftmargin=*]
  \item We conduct a systematic analysis of the alignment issues in text-to-image diffusion models, which occur not only on the text encoding side but also on the image side and the cross-attention blocks.
  \item We propose a latent object detection method that fully utilizes diffusion model alignment information. Our detection model, trained on COCO, exhibits good generalization to unseen categories.
  \item We introduce Detector Guidance to address the weak global comprehension of diffusion models, which provides a significant advantage without any human involvement or additional iterations.
\end{itemize}

\section{Related Work}

In this section, we introduce diffusion models and focus on Stable Diffusion, the basis of our method.

\subsection{Diffusion Model} 

The diffusion denoising models \citep{sohl2015deep,ho2020denoising,song2021scorebased} are a type of deep generative models that employ an iterative denoising process to generate samples. These models utilize noise-conditioned score networks, as described in \cite{song2019generative,song2020improved}, and denoising score matching objectives, as described in \cite{hyvarinen2005estimation,vincent2011connection}, at varying noise levels. They have demonstrated successful applications in various domains, such as text-to-image generation \cite{saharia2022photorealistic,ramesh2022hierarchical,rombach2022high}, natural language generation \cite{li2022diffusion}, time series prediction \cite{tashiro2021csdi}, audio synthesis \cite{kong2020diffwave, liu2022diffsinger}, 3D point cloud generation \cite{luo2021diffusion}, and molecular conformation generation \cite{xu2022geodiff}.

\textbf{Text-to-Image Generation} Diffusion models play an important role in text-to-image generation. To improve computational efficiency, diffusion models are typically trained on low-resolution images \citep{saharia2022photorealistic} or latent variables \citep{rombach2022high, gu2022vector}, which are then transformed into high-resolution images through super-resolution diffusion models \cite{ho2022cascaded} or latent-to-image decoders \cite{sinha2021d2c}. The sampling process of diffusion models utilizes classifier-free guidance \cite{ho2022classifier} as well as various sampling algorithms that use deterministic \cite{song2021denoising,lu2022dpm,liu2022pseudo,zhang2022fast,karras2022elucidating} or stochastic \cite{bao2022analytic, dockhorn2021score,zhang2022gddim} iterations. In addition, several works \cite{blattmann2022retrieval,sheynin2022knn} retrieve additional images related to the text prompt from an external database and use them to condition generation, thereby enhancing performance.

\textbf{Multi-Object Generation} Multi-object generation needs text-to-image models to comprehend different objects in the generation process. As the difficulty in the alignment between text and image, prior studies have utilized bounding boxes \citep{ma2023directed}, masks \citep{avrahami2022spatext}, or small patches \citep{sarukkai2023collage} of target objects to enhance alignment. Some studies aim to improve generation results without the need for human involvement in each generation, \citet{liu2022compositional} proposed an approach where concept conjunctions are achieved by adding multiple estimated scores for different objects. And \citet{feng2022training} utilize language parsers to associate attributes solely with the corresponding objects. \citet{balaji2022ediffi} combine T5 \citep{raffel2020exploring} and CLIP \citep{radford2021learning, cherti2022reproducible} text encoders to improve the alignment in the text side.

\textbf{Applications} The diffusion models for text-to-image have a significant impact on downstream applications. These models can be directly applied to various inverse problems, such as super-resolution \cite{saharia2022image, dhariwal2021diffusion}, inpainting \cite{lugmayr2022repaint,chung2022improving}, and JPEG restoration \cite{saharia2022palette,kawar2022jpeg}. Text-to-image diffusion models can also be used for other semantic image editing tasks. For instance, SDEdit \cite{meng2021sdedit} enables editing of an existing image via colored strokes or image patches. DreamBooth \cite{ruiz2022dreambooth} and Textual Inversion \cite{gal2022image} allow for personalized model implementation by learning a subject-specific token from a few images. Similar image-editing capabilities can also be achieved by fine-tuning the model parameters \cite{kawar2022imagic,valevski2022unitune} or automatically searching for editing masks using denoisers \cite{couairon2022diffedit}. Several text-to-video diffusion models are developed on text-to-image ones and achieve high-quality video generation results \cite{ho2022video,ho2022imagen,yang2022diffusion,singer2022make,harvey2022flexible}. Furthermore, the fitness capabilities of diffusion models have proven beneficial for the task of out-of-distribution detection \cite{liu2022diffusion}.

\subsection{Stable Diffusion}

Our method is constructed on a SOTA open-source text-to-image model: Stable diffusion, which comprises an autoencoder and a diffusion model. During the training process, the pre-trained autoencoder first compresses the image distribution into a latent space, and the diffusion model attempts to fit this new distribution in the latent space. In the sampling process, the diffusion model first generates latent results based on the text prompts and then uses the autoencoder to decode the latent results and obtain the final images.

Stable Diffusion utilizes a pre-trained CLIP model to encode the text prompt and incorporates multiple cross-attention blocks to integrate text information into the target regions of images. Previous work \citep{hertz2022prompt} has shown that these cross-attention blocks contain rich spatial structure information and control the spatial layouts of the generated results. This observation provides us with the possibility of using such spatial information to separate different objects and correcting the following cross-attention blocks to solve the mixing problems.

\section{Detector Guidance}

In this section, we begin by discussing the structural information that can be obtained from the cross-attention maps (CAMs) of diffusion models and explain why additional models are necessary to facilitate generation. We then illustrate how a latent object detection model can be integrated into the diffusion models. Subsequently, we combine the results of bounding box detection and CAMs to achieve multiple object segmentation in a noisy latent space. We utilize the segmentation results to eliminate conflicting information and enhance target information by refining the subsequent CAMs. Finally, we present our detector guidance method that incorporates all of these components.

\begin{figure}
  \centering
  \includegraphics[width=0.9\linewidth, trim=0 5 0 0]{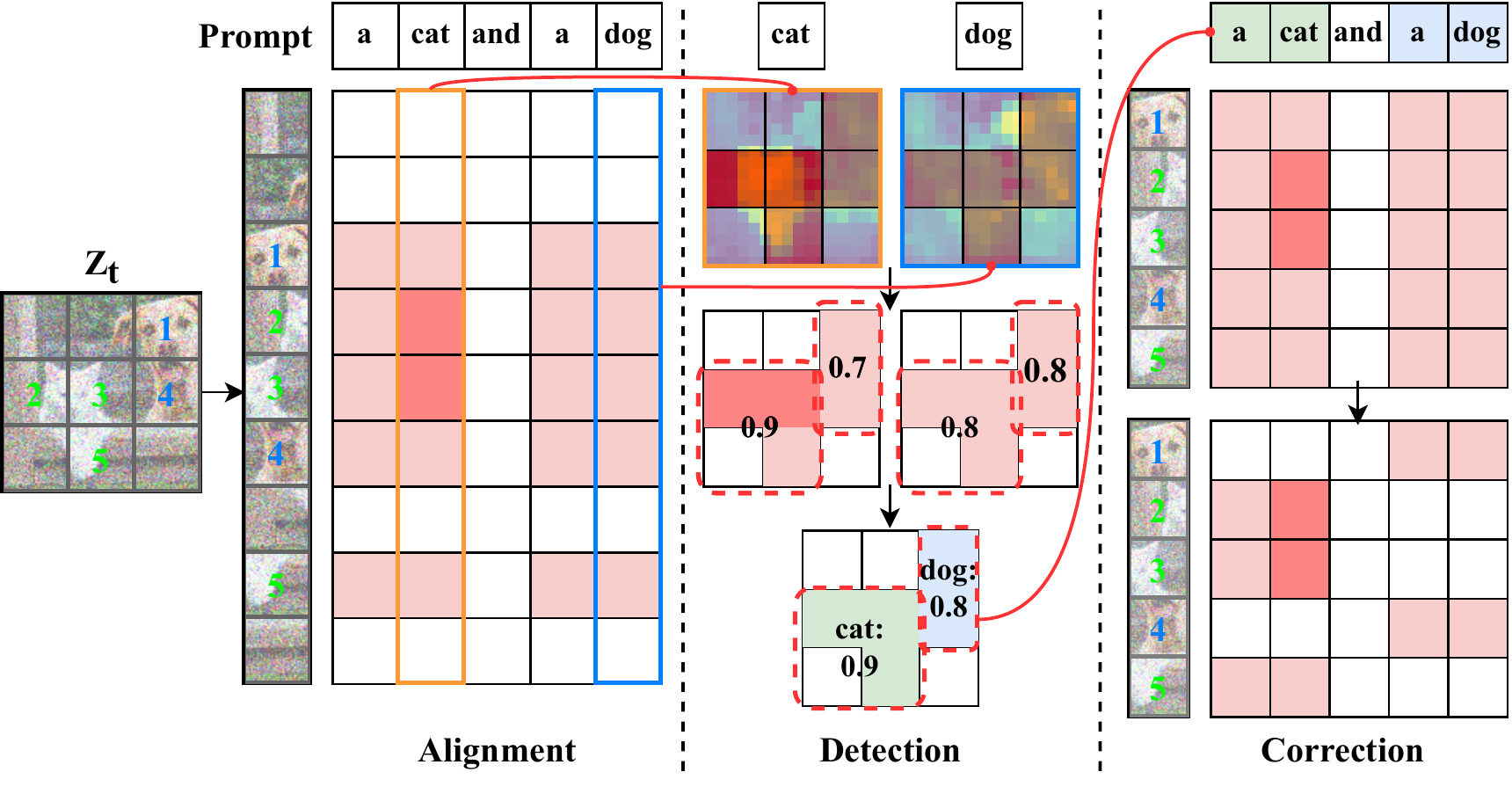}
  \caption{The pipeline of Detector Guidance comprises two stages: detection and correction. In the detection stage, CAMs of nouns are resized into squares, and latent object detection is performed on them. We then assign different regions to different objects to maximize the total confidence value. In the correction stage, we utilize the global alignment between regions and prompts by masking conflicting alignment and enhancing target alignment to generate new, corrected CAMs.}
  \vspace*{-0.3cm}
\end{figure}

\subsection{Cross-Attention Map}

The multiple cross-attention blocks in Stable Diffusion exhibit remarkable alignment abilities in text-to-image generation. In a cross-attention block, the features $z_t$ of the noisy data are projected to a key $K=L_K(z_t)$, and the textual embedding $c$ is projected to a query $Q=L_Q(c)$ and a value $V=L_V(c)$, via learned linear projections. The cross-attention map (CAM) is then $M = KQ^T$, and the final cross-attention output of this block is $\text{Softmax}(M/\sqrt{d})V$. Here, $d$ is the projection dimension of the key and query. Among multiple CAMs at different resolutions, CAMs in the middle have rich spatial structure information \citep{hertz2022prompt}. Spatial information can still be clearly observed even when only 20\% of the generation process has been completed, as shown in Figure \ref{fig_cam}.

While such alignment is effective in many cases, it lacks global coordination, leading to disorderly competition. For instance, consider a text prompt ``a leopard and a tiger'' and the intermediate results correctly generate two objects. Ideally, the leopard prompt should align with one region, and the tiger prompt should align with another. However, in practice, the leopard and tiger prompts may both attempt to align with two regions simultaneously, resulting in the mixing of conflicting information and the generation of leopard and tiger hybrids, as shown in Figure \ref{fig_cam}. Moreover, designing global coordination solely with a pre-trained diffusion model is challenging due to its local alignment strategy, which results in limited global comprehension abilities for distinguishing between different objects. Thus, an additional model is necessary to identify objects from the local alignment information of CAMs, which we refer to as latent object detection. %

Although we incorporate an additional model, our latent object detection model can be simple for several reasons. Firstly, local alignment has already marked important areas, making it easier to extract features. Secondly, the resolution of the middle CAMs is only $16\times16$, resulting in a relatively small scale difference between objects. Thirdly, we typically avoid images where objects overlap each other, as our objective is to present the objects mentioned in the text as comprehensively as possible. These factors make the detection process relatively straightforward.

\subsection{Latent Object Detection}

In this paper, we use a simple YOLO \citep{redmon2016you} model for latent object detection and train it on the COCO dataset. Our training procedure is as follows: we add Gaussian noise into the images and use the labels of the original images as the prompts. We then feed the noisy images and prompts into a pre-trained diffusion model. We capture the outcomes of middle CAMs, which we subsequently employ as input to the latent object detection model. The latent object detection model infers the bounding boxes and confidence scores for each object based on the corresponding CAMs. We then use the predicted bounding boxes and the ground truth bounding boxes to calculate the loss outcomes and update the latent object detection model correspondingly. More details and analyses about our latent detection model can be found in Appendix \ref{app_lod}. 

In the sampling procedure, we first utilize a language parser \citep{bird2009natural} to find the objects in prompts. More details about language parsers can be found in Appendix \ref{app_parser}. Then we generate bounding boxes for each object based on the corresponding CAMs, with the class for each bounding box being the noun corresponding to the input CAM. The next step is to assign bounding boxes to different objects within a single image. To achieve this, we first employ non-maximum suppression to eliminate redundant bounding boxes. We do not discard all information from unused bounding boxes. Rather, we incorporate the confidence score and object class of unused bounding boxes into the corresponding choice bounding boxes. As a result, each choice bounding box may pertain to different objects with varying scores. We employ the linear sum assignment algorithm from the SciPy library to assign bounding boxes to different objects and optimize the total confidence score across all objects.

Despite the relatively limited number of categories in the COCO dataset compared to larger-scale datasets like Laion-5B \citep{schuhmann2022laion}, we find that our well-trained detection model demonstrates good generalization to previously unseen categories. The use of CAMs as input has played a significant role in this. Many previously unseen categories, which may exhibit substantial visual differences at the image level, are remarkably similar to certain known categories in terms of their CAMs as long as they share similar overall structures. %

\subsection{CAM Correction}

After obtaining the desired bounding boxes for each object, we use three consecutive steps to correct the CAMs and a strategy to maintain continuity in the generation process.

\textbf{Boundary Correction} Theoretically, we can use segmentation models instead of detection models to obtain more precise object boundaries. However, we find that this is often unnecessary, as the CAMs already contain sufficient local boundary information for objects. Therefore, we first use the detection model to generate bounding boxes $\text{BBox}[*]$ and then further refine the boundaries using CAMs. The segmentation for a certain object $n$ is:
\begin{equation}
  S[n](i,j) = \{(i,j) \in \text{BBox}[n]\text{ \& }\text{CAM}[..., n](i, j) >= \sigma[n]\}.
\end{equation}
Here, the shape of CAM is $H\times W \times N$, and $N$ is the total number of tokens in a prompt. We only do detection and boundary correction on the nouns that represent objects, and the shape of a segmentation of object $n$ is $H\times W$. The threshold $\sigma$ is computed using Otsu's method \citep{otsu1979threshold}. For pixels belonging to more than one object, we assign them to the smallest object. %

\textbf{Conflict Elimination} We find that the alignment results in CAMs become meaningful after $t\leq 800$.\footnote{Visual results for the meaningful CAMs be found in Section \ref{sec_visual}. The transition timestep is $I=800$.} Then, we can address the issue of information mixing. In each cross-attention block, we utilize the conditional prompt $c$ and the unconditional prompt $uc$ to generate two queries $Q_c$, $Q_{uc}$ and two CAMs $KQ_c^T$, $KQ_{uc}^T$, respectively. Here, we denote $\text{CAM}_0=KQ_c^T$. For any region that has established its correspondence with the text describing a certain object, we eliminate the influence of other conflicting texts by replacing the values of the corresponding $KQ_c^T$ with those of $KQ_{uc}^T$ at the same position. Conflicting relationships can be obtained through human annotations or a language parser. Therefore, the corresponding mask is:  
\begin{equation}
  \text{mask}(i,j,p)=\{\exists n\text{ s.t. }S[n](i,j)=1\text{ \& }p \text{ conflicts with } n\}.
\end{equation}  
Here, the shape of the mask is identical to that of a CAM. Other tokens in a noun phrase share the same mask as the core noun. The Conflict Elimination algorithm can be expressed as follows:
\begin{equation}
  \text{CAM}_1=\text{CAM}_0 * (1 - \text{mask}) + KQ_{uc}^T * \text{mask}.
\end{equation}

\textbf{Target Enhancement} While the masking operator can prevent conflicting information from being mixed in the subsequent steps, it does not guarantee that the correct information can have sufficient influence in the target region to generate the right objects. One reason is that some mistakes may have already occurred in the previous steps, which can affect the alignment between pixels and target text. To address this issue, we propose Target Enhancement that strengthens the influence of the target text in such regions. We record the maximum value of CAMs for each pixel. If the maximum decrease after Conflict Elimination, it indicates that this pixel contains a large percentage of features belonging to other objects. In this case, we increase the value of the remaining CAMs at this pixel to enhance the injection of target information. The algorithm of Target Enhancement can be written as:  %
\begin{equation}
  \text{CAM}_2 = \text{CAM}_1 * \frac{\text{CAM}_0.\text{max}(\text{dim}=-1)}{\text{CAM}_1.\text{max}(\text{dim}=-1)}.
\end{equation}

\textbf{Smooth Involvement} Another issue with masking operators is that they can cause a sharp change in the outputs when first applied. Such discontinuous outputs are not suitable for high-order numerical acceleration methods of diffusion models, such as PNDM \citep{liu2022pseudo}, DPM-Solver \citep{lu2022dpm}, and DEIS \citep{zhang2022fast}. To address this, we propose a smoothing approach that gradually increases the impact of Detector Guidance. Specifically, we begin by using 25\% $\text{CAM}_2$ and 75\% original CAM and gradually increase the ratio to 50\%:50\%, 75\%:25\%, and 100\%:0\%.  %

Upon the introduction of all the steps, the whole algorithm of our Detector Guidance can be found in Algorithm \ref{alg_dg_d} and \ref{alg_dg_c}. A more comprehensive version can be found in Appendix \ref{app_alg}. Notably, our CAM correction has no hyperparameters for different situations, making it easy and robust to use. We also find the bounding boxes can be cached and reused in several successive steps without affecting alignment and quality, thereby improving the computational efficiency of our method.

\begin{figure}[t]
  \begin{minipage}{.49\linewidth}
    \begin{algorithm}[H]
      \caption{Detection of Detector Guidance}
      \label{alg_dg_d}
      \begin{algorithmic}[1]
        \Require initial noise $x_T$, text condition $c$
        \For {$t = T, T-1, \cdots, T - I + 1$}
          \State $x_{t-1}$, $\text{CAM}$ = DM($x_t$, $c$)
        \EndFor 
        \For {$t = T - I, T- I-1, \cdots, 1$}
          \State $x_{t-1}$, $\text{CAM}$ = DM($x_t$, $c$, $\text{BBox}_{t+1}$, $\sigma_{t+1}$)
          \State $\text{BBox}_t$ = Detection($\text{CAM}$)
          \State $\sigma_t$ = Otsu($\text{CAM}$)
        \EndFor \\
        \Return image = Decoder($x_0$)
      \end{algorithmic}
    \end{algorithm}
  \end{minipage}
  \ \ 
  \begin{minipage}{.49\linewidth}
    \begin{algorithm}[H]
      \caption{Correction of Detector Guidance}
      \label{alg_dg_c}
      \begin{algorithmic}[1] %
        \Require bounding box $\text{BBox}_{t+1}$, threshold $\sigma_{t+1}$, smoothing factor $s$
        \For {block $\in$ cross-attention blocks}
          \State $\text{CAM}_0$, $\text{CAM}_{\text{uc}}$ = $KQ_{c}^T$, $KQ_{uc}^T$
          \State $\text{m}$ = Mask($\text{CAM}_0$, $\text{BBox}_{t+1}$, $\sigma_{t+1}$)
          \State $\text{CAM}_1$ = $\text{CAM}_0 * $(1 - $\text{m}$) + $\text{CAM}_{\text{uc}} * \text{m}$
          \State $\text{CAM}_2 = \text{CAM}_1 * \frac{\text{CAM}_0.\text{max}(\text{dim}=-1)}{\text{CAM}_1.\text{max}(\text{dim}=-1)}$
          \State $\text{CAM}_3$ = $\text{CAM}_2 * s$ + $\text{CAM}_0 * (1 - s)$
          \State output = $\text{Softmax}(\text{CAM}_3/\sqrt{d}) * $ V
        \EndFor
      \end{algorithmic}
    \end{algorithm}
  \end{minipage}
  \vspace*{-0.2cm}
\end{figure}

\section{Experiments}

In this section, we present the setups of our method and compare it with the baselines on COCO, CC, and a new benchmark MRO. Then we showcase the performance of our latent object detection model at various timesteps and the influence of modifying CAMs using Detector Guidance. After that, we conducted ablation studies to analyze the effectiveness of each step in Detector Guidance.

\subsection{Setups}

We evaluate Detector Guidance on pre-trained Stable Diffusion models v1.4 and v2.1. The additional latent detection model is YOLOv1 with the conv2d stride of 1 to suit a small input size. The latent detection model is trained on the COCO dataset, utilizing 2 RTX3090*days, with nearly 70\% of the time allocated to CAMs computation.

Here, we evaluate the alignment of our DG method on three benchmarks. First is the Concept Conjunction (CC) benchmark from \citep{feng2022training}, which comprises about 500 prompts following the structure of ``a red car and a white sheep''. This is a suitable benchmark for Stable Diffusion v1.4 but too simplistic for Stable Diffusion v2.1. Consequently, we present a novel benchmark, the multi-related object (MRO) benchmark, which uses a similar sentence pattern. However, instead of using distinct objects such as a car and a sheep, we utilize GPT4 \citep{openai2023gpt4} to generate 30 prompts that contain two related objects, such as a tiger and a leopard. Additionally, we generate 10 samples for each prompt instead of just 1. This benchmark evaluates the ability of generation models to solve mixing problems both at the attribute level and at the object level. The complete MRO list can be found in Appendix \ref{app_mro}. Moreover, the COCO validation set is a commonly used benchmark for zero-shot text-to-image generation. We utilize it to evaluate the performance under more complex prompt patterns.

Regarding evaluation metrics, we find that FID and CLIP-score are not effective indicators of the mixing problem. Therefore, we primarily rely on human evaluation. Nevertheless, we also evaluate FID and CLIP-score to ensure that our method does not lead to any performance degradation on these two metrics. In Figure \ref{fig_m+c}, we also present the visual generation results on MRO and COCO.

\subsection{Main Results}

\begin{figure}[t]
  \begin{minipage}[t]{0.34\linewidth}
    \centering
    \includegraphics[width=\linewidth, trim=5 5 5 5]{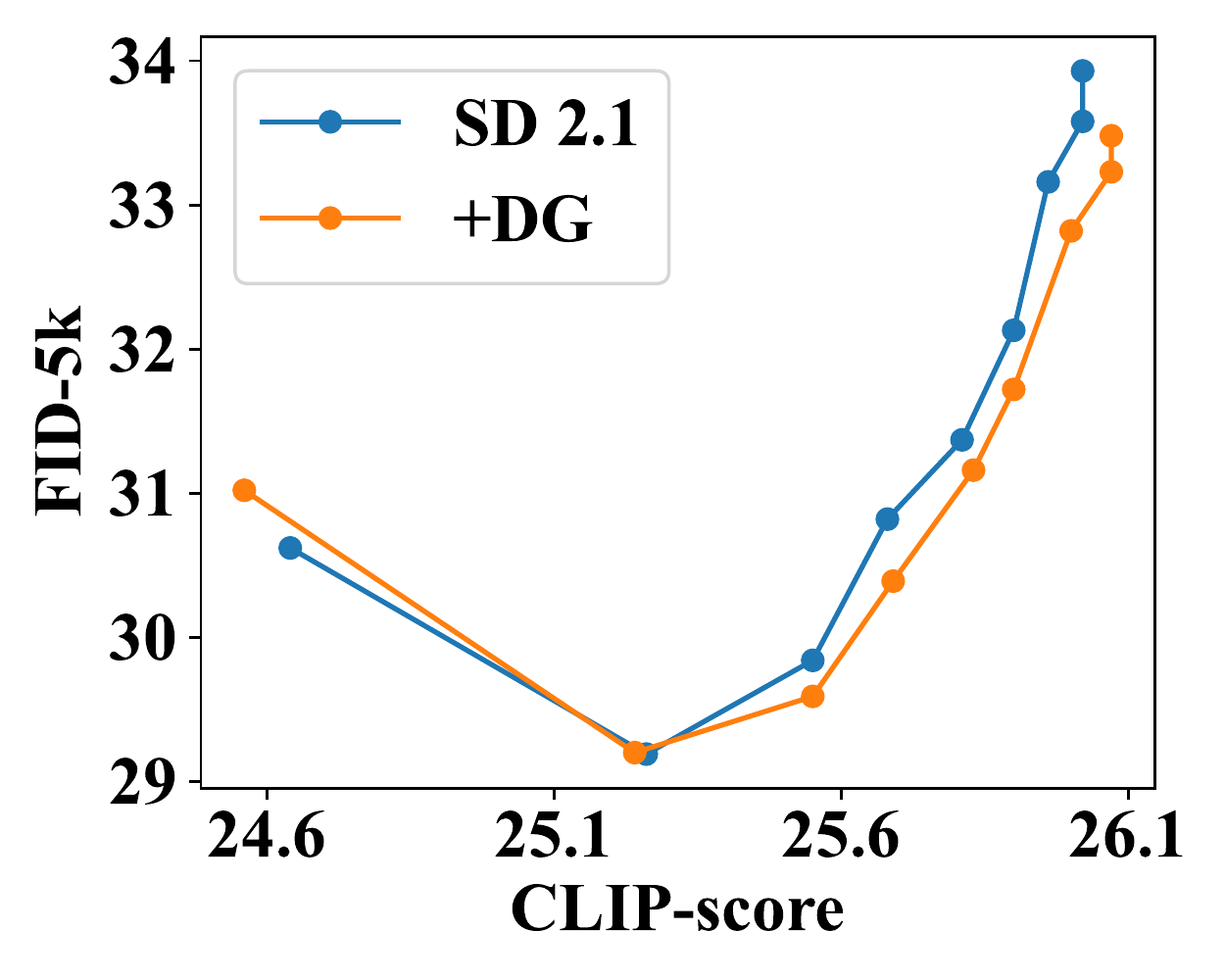}
    \vspace*{-0.55cm}
    \caption{The trade-off curve between FID-5k and CLIP-score for guidance scales of $[2.0, \cdots, 10.0]$.}
    \label{fig_curve}
    \vspace*{-0.2cm}
  \end{minipage}
  \ \ 
  \begin{minipage}[t]{.64\linewidth}
      \vspace*{-3.72cm}
      \centering
      \small
      \setlength\tabcolsep{4pt}
      \begin{tabular}{@{}lllccc@{}}
        \toprule
        \multirow{2}{*}{Model} & \multirow{2}{*}{Guidance} & \multirow{2}{*}{Benchmark} & \multicolumn{3}{c}{Who aligns better?} \\ \cmidrule(l){4-6} 
        &  &  & Base & Guidance & Delta \\ \midrule
        \multirow{3}{*}{SD 1.4} & Structure \citep{feng2022training} & \multirow{3}{*}{CC} & 27.4\% & 29.4\% & +2.0\% \\
        & DG (ours) &  & 15.9\% & 23.5\% & +7.6\% \\
        & \small{Str+DG (ours)} &  & 24.4\% & 34.3\% & \textbf{+9.9\%} \\ \midrule
        \multirow{3}{*}{SD 2.1} & \multirow{3}{*}{DG (ours)} & CC & 14.3\% & 15.0\% & +0.7\% \\
        &  & MRO & 13.0\% & 35.3\% & \textbf{+22\%} \\
        &  & COCO & 12.8\% & 23.6\% & +11\% \\ \bottomrule
      \end{tabular}
      \caption{The human evaluation results of the alignment between text and image. Except for cases where the base or guidance method is better, the remaining results are tied.}
      \label{tab_cc_mro}
  \end{minipage}
\end{figure}

\begin{figure}[t]
  \vspace*{0.1cm}
  \centering
  \includegraphics[width=\linewidth, trim=0 5 0 0]{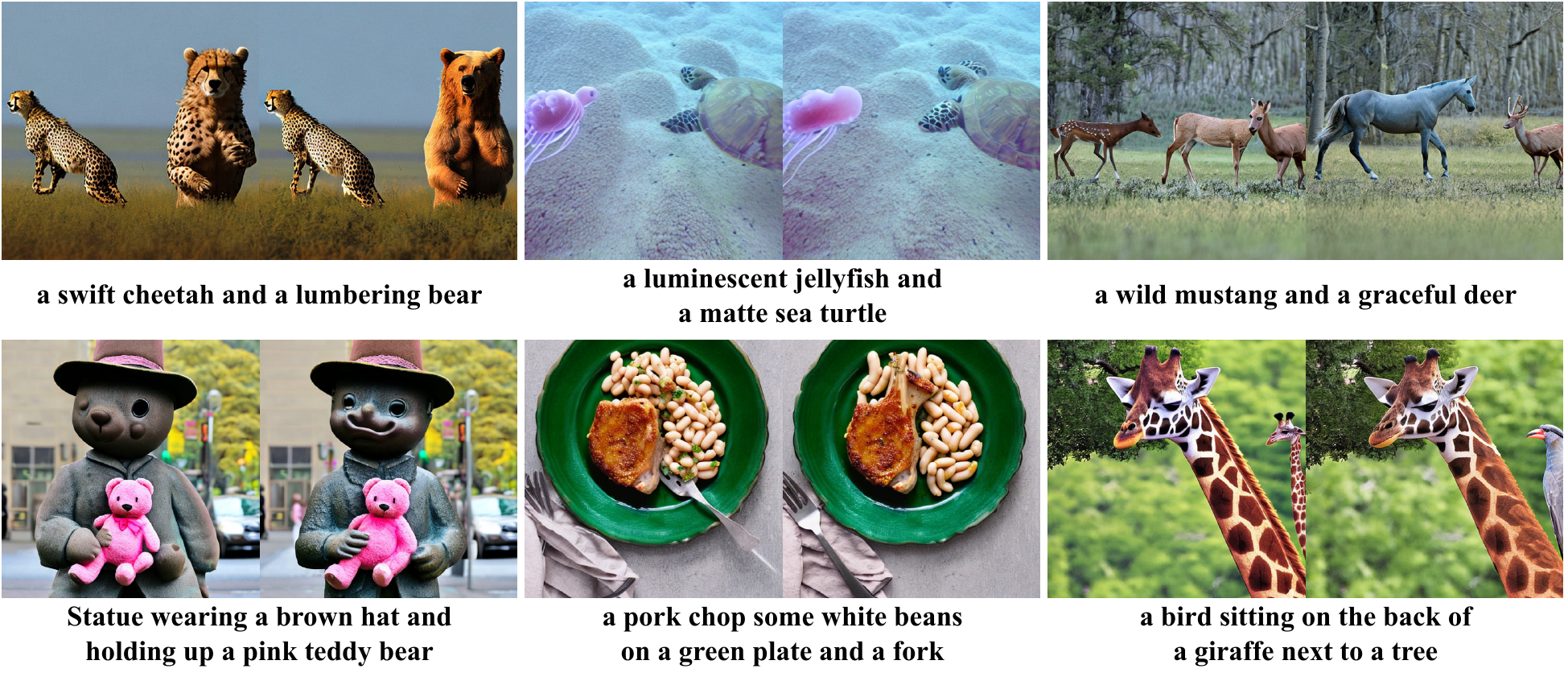}
  \caption{Each pair of images compares Stable Diffusion v2.1 without (left) or with (right) DG. The first line shows DG's effectiveness for attribute mixing, object mixing, and object disappearance, while the second line demonstrates its effectiveness with complex prompts from COCO.}
  \label{fig_m+c}
  \vspace*{-0.4cm}
\end{figure}

\textbf{CC} Structured diffusion guidance \citep{feng2022training} is another guidance method that focuses on the text side to address the mixing problem. This work is built on Stable Diffusion v1.4 and provides the CC benchmark. Thus, we compare it with our method on the same benchmark and model. Moreover, since the two methods address the mixing problem on the text and image sides, respectively, they can be combined. In Table \ref{tab_cc_mro}, we find that the combination of our method and the structured diffusion guidance method can further enhance the performance. One issue with structured diffusion guidance is that it may introduce unnecessary adjustments and is more likely to result in guidance results that are worse than the original ones, about 10\% worse than ours.

\textbf{MRO} We use the more challenging MRO benchmark to evaluate the performance of our method on Stable Diffusion v2.1. The results are shown in Table \ref{tab_cc_mro}. We can see that our method still achieves huge improvements on Stable Diffusion v2.1. As shown in the first row of Figure \ref{fig_m+c}, Detector Guidance successfully solves the attribute mixing, object mixing, and object disappearance problems.

\textbf{COCO} In our experiments, we randomly select 5,000 captions from all COCO captions and generate 5,000 corresponding image pairs with and without Detector Guidance. To perform latent object detection on COCO, we use a language parser to identify all the noun phrases within the captions from COCO. We plot the trade-off curves between FID and CLIP-score using different guidance scales, including $[2.0, 3.0, \cdots, 10.0]$. Figure \ref{fig_curve} shows that Detector Guidance improves a bit the results in both FID and CLIP-score when the guidance scale is larger than 3. To evaluate our method under complex prompt patterns, we conducted human evaluations on 500 pairs of images with the largest L2 distance from the 5,000 pairs. The results also show a clear advantage of our method.

\subsection{Analyses}
\label{sec_visual}

\begin{figure}[t]
  \begin{minipage}[t]{0.37\linewidth}
    \includegraphics[width=\linewidth, trim=5 5 5 5]{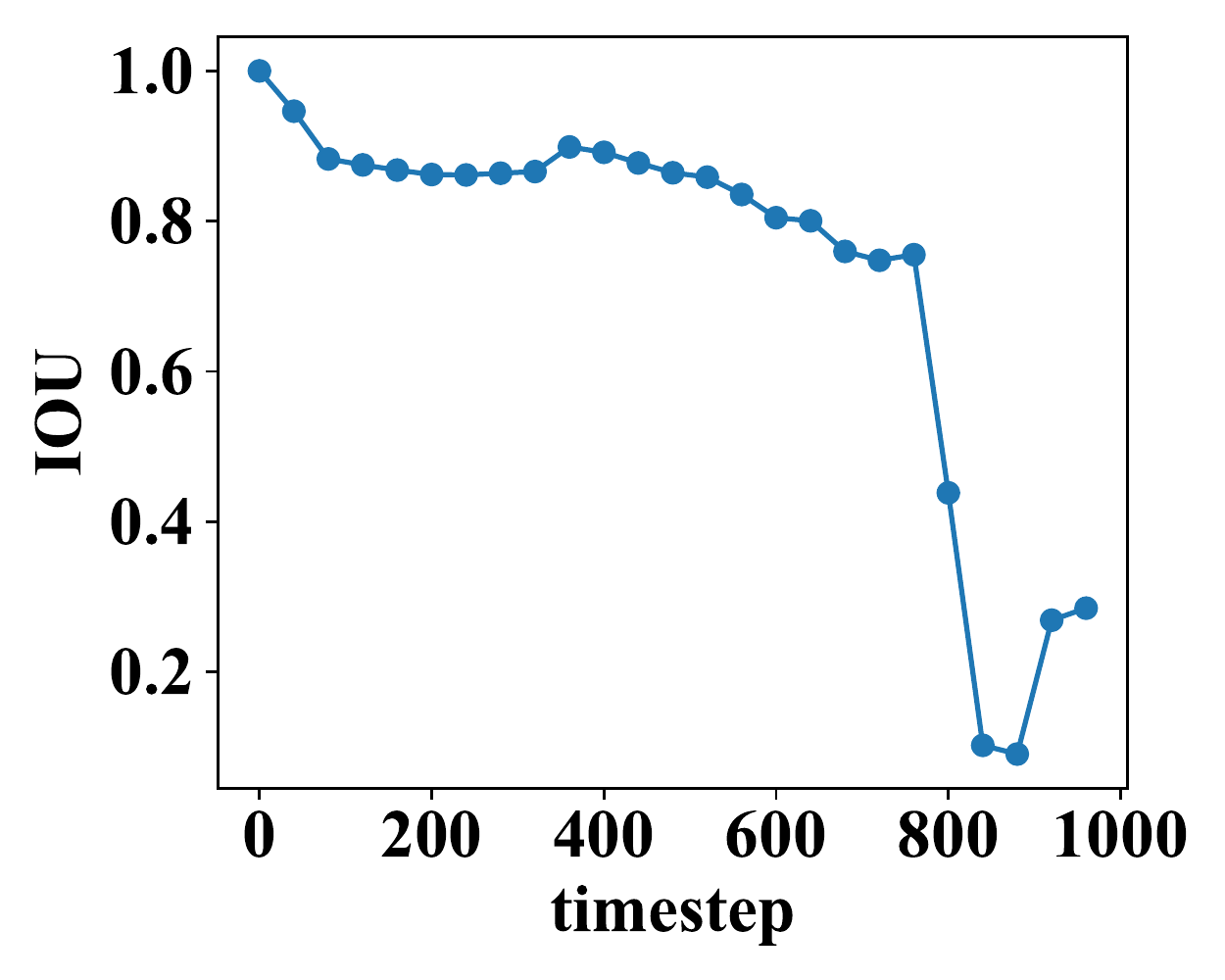}
    \caption{The IOU between the predicted bounding boxes at different timesteps and at the last step.}
    \label{fig_iou}
  \end{minipage}
  \ \ \ 
  \begin{minipage}[t]{0.60\linewidth}
    \includegraphics[width=\linewidth, trim=15 0 0 5]{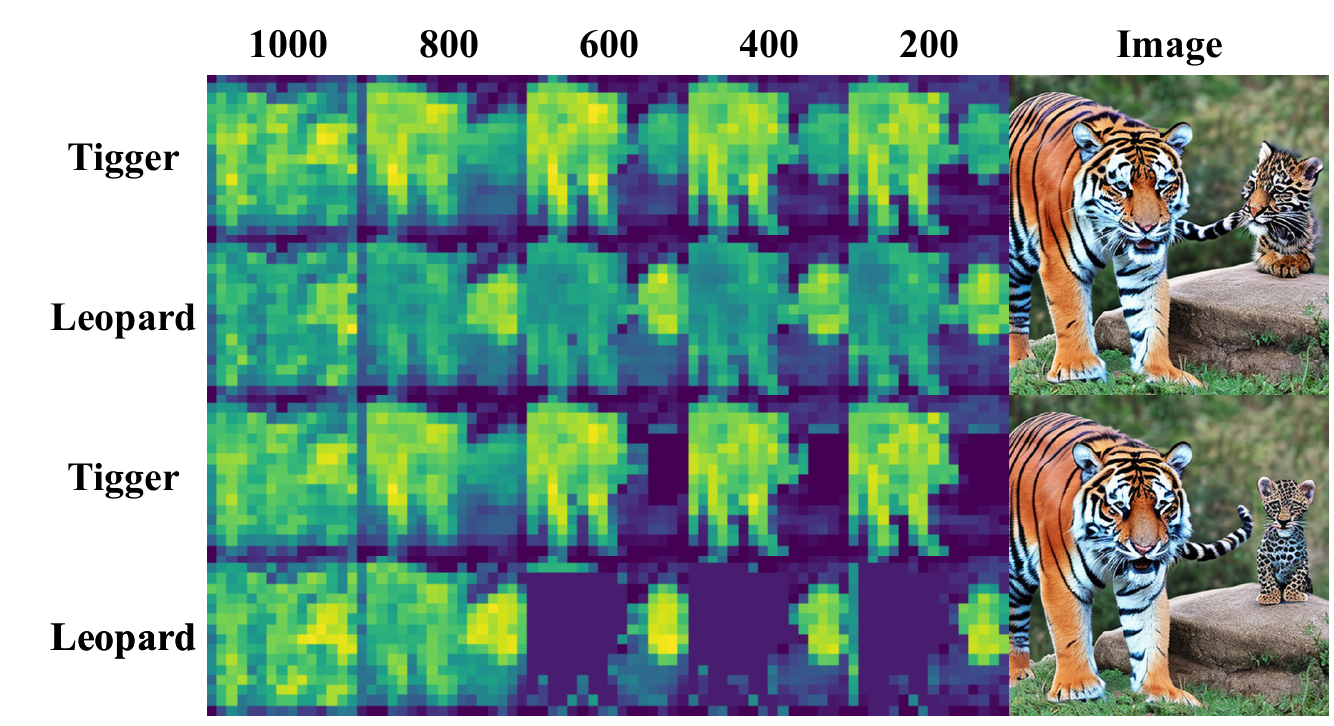}
    \caption{The CAMs at different timesteps with or without DG. The prompt is ``A striped tiger and a spotted leopard''. The initial noise is the same.}
    \label{fig_cam}
  \end{minipage}
  \vspace*{-0.2cm}
\end{figure}

\textbf{Detection} Based on our observations, we notice that independent objects can be discerned from CAMs when $t \leq 800$. This is in strong agreement with the outcomes we obtained using a latent object detection model that was well-trained on CAMs. In Figure \ref{fig_iou}, we find that the IOU results undergo significant changes when $t \geq 800$ and gradually converge to one when $t \leq 800$. Therefore, it can be deduced that for $t \geq 800$, the latent object detection model essentially makes random guesses regarding the object, while for $t \leq 800$, the model can effectively locate the object and subsequently refine its boundaries, ultimately converging to the final result.

Our detection model is only trained on the COCO dataset. Nonetheless, we observe that it exhibits good generalization to unseen categories. Notably, most objects in the demos of this paper do not belong to any category included in COCO. Our method can still detect objects and align objects with these prompts successfully. %

\textbf{Correction} In Figure \ref{fig_cam}, we show an example of CAMs associated with the tokens ``tiger'' and ``leopard'' under different timesteps with or without DG. Here, the whole prompt is ``A striped tiger and a spotted leopard''. In the original diffusion model, the CAMs attempt to align the tokens ``tiger'' and ``leopard'' with both objects simultaneously. Consequently, the information from ``tiger'' and ``leopard'' amalgamates in the right object, resulting in an animal that resembles both a tiger and a leopard. Upon incorporating DG into the diffusion model, DG can accurately distinguish between the two objects and align each one with its corresponding prompt, resulting in a final result that is highly consistent with the conditions. What's more, DG adheres to the principle of minimal intervention, as it successfully preserves the image elements, such as rocks, vegetation, and the tiger, except for the region where the mixing error occurred. This reduces the possibility of introducing unknown new problems through additional corrections.

\subsection{Ablation Studies}

\begin{wrapfigure}{r}{0.6\textwidth}
  \centering
  \includegraphics[width=\linewidth, trim=0 10 0 0]{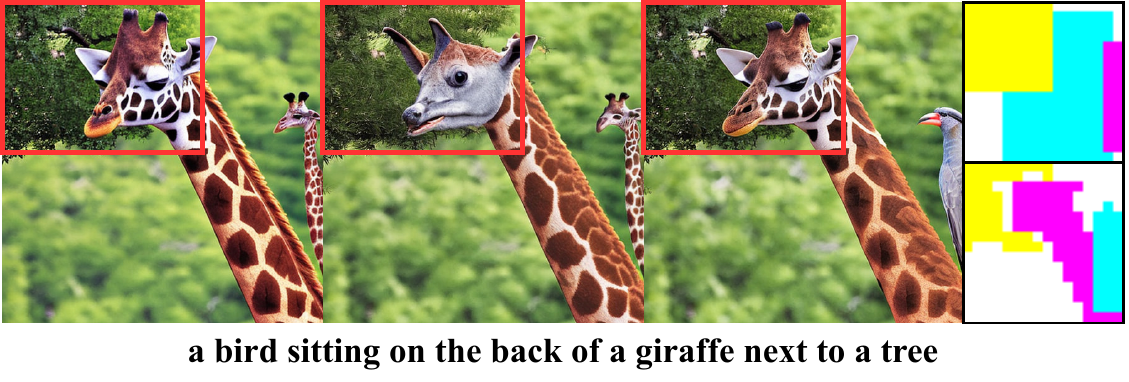}
  \caption{The results with or without Boundary Correction. The bounding box corresponds to the object ``tree'', and the last two images are masks.}
  \label{fig_refine}
  \vspace*{0.1cm}
  \includegraphics[width=\linewidth, trim=0 10 0 0]{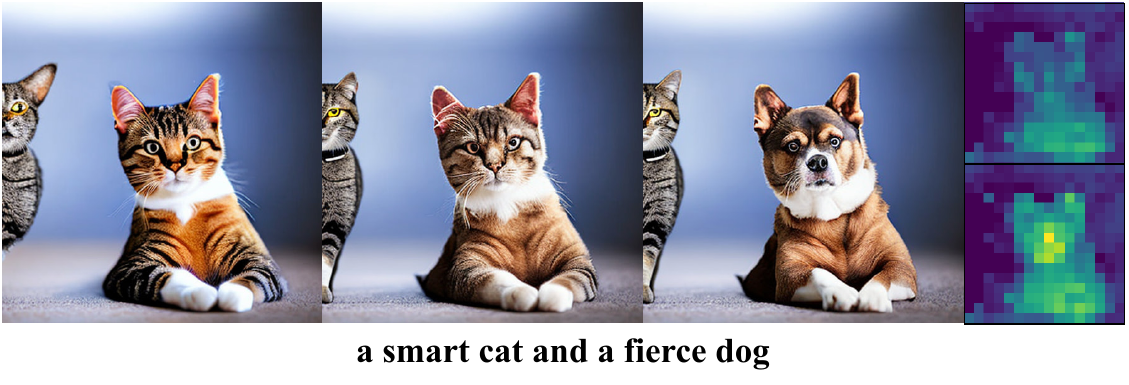}
  \caption{The image results and CAMs of the token ``dog'' with or without Target Enhancement.}
  \label{fig_enhance}
  \vspace*{-0.8cm}
\end{wrapfigure}

\textbf{Boundary Correction} Since bounding boxes of different objects may overlap with each other, some areas of one object may be mistakenly assigned to other objects. Boundary Correction can address this issue, as illustrated in Figure \ref{fig_refine}. In the second image, the boundary box of the tree overlaps with the boundary box of the giraffe, leading to poor generation results. Boundary Correction can refine
the boundary of the tree and remove the overlapping area. As a result, the hand of the giraffe can be generated correctly in the third image.

\textbf{Target Enhancement} While Conflict Elimination cannot guarantee proper alignment of the target prompt with the assigned area, we use Target Enhancement to enhance the target object, as shown in Figure \ref{fig_enhance}. In the first image, both the left and right sides of the image bear a resemblance to a cat. Without Target Enhancement, even though we remove the cat prompt on the right and improve the left cat in the second image, the dog prompt cannot effectively align the right region. In the third image, Target Enhancement resolves this issue by enhancing the target object in CAMs.

\textbf{Smooth Involvement} Smooth Involvement is an important theoretical issue, but we did not observe significant improvement in practice. A similar phenomenon occurs in the acceleration of diffusion models, which use lower-order methods to start the generation and still yield good generation results.

\section{Discussion}

In this paper, we introduce Detector Guidance to aid diffusion models in comprehending global object information. In the detection stage, we do not perform detection directly on the image space. Instead, we utilize the local alignment information from Stable Diffusion and latent object detection to obtain robust detection results. In the correction stage, we adhere to the principle of minimal intervention and avoid the use of any hyperparameters to ensure the easy, robust, and safe application of our method. Additionally, we provide a new benchmark to assess the performance of generation models on the multi-object mixing problem. 

Apart from text-to-image generation, Detector Guidance can be effortlessly implemented in other diffusion models that also utilize cross-attention blocks and encounter the problem of multi-object mixing. For instance, Imagebind \citep{girdhar2023imagebind} achieves alignment of more modalities and images, such as text, audio, depth, and thermal. The problem of audio guidance being from different objects is similar to that of text prompts containing multiple objects. We believe that all of these modalities can be utilized as guidance in the future and can benefit from our Detector Guidance.

{
  \small
  \bibliography{neurips_2023}
  \bibliographystyle{unsrtnat}
}

\newpage

\appendix

\section{Supplementary Material}

\subsection{Latent Object Detection}
\label{app_lod}

Our latent object detection is built on YOLOv1 and Stable Diffusion v2.1. Specifically, we adopt an implementation of YOLOv1 from \url{https://github.com/yjh0410/new-YOLOv1_PyTorch}. More details are as follows:
\begin{itemize}[leftmargin=*]
  \item \textbf{Input}: The input is the concatenation of 5 CAMs in the Unet with a size of $20 \times 16 \times 16$ corresponding to the core noun, and we only use the mean CAMs along the first head dimension. If the core noun consists of multiple tokens, we use the average CAMs across all these tokens, too.
  \item \textbf{Model Structure}: To suit a small input size, we employ ResNet as the backbone with a conv2d stride of 1 and use several conv2d layers as the only head to simultaneously predict the bounding box and confidence.
  \item \textbf{Loss Function}: We exclude the classification loss and only retain the confidence loss and txtytwth loss from YOLOv1.
  \item \textbf{Dataset}: We utilize the COCO dataset, and adopt the official train/val split. We then use the validation set as the test set.
  \item \textbf{Augmentation}: We employ several augmentations, including random horizontal flipping, random brightness adjustment, and random cropping.
  \item \textbf{Optimizer}: We use the AdamW optimizer with a learning rate of 0.001 and a batch size of 32. We train the model for 100k steps and use the final model.
  \item \textbf{Evaluation Metric}: We use the same evaluation metric as YOLOv1, which is the mean average precision (mAP) with an IoU threshold of 0.5. 
\end{itemize}

Table \ref{tab_lod} presents the results, which demonstrate that our latent object detection model achieves a mAP$_{.5}$ of 58.4. This confirms that the CAMs contain rich semantic information, and indicates that our approach can effectively detect objects in the latent space. As the noise level increases, the detection results decrease. This is because the noise confuses the model, leading to a decline in detection performance.

\begin{table}[h]
  \vspace*{-0.3cm}
  \centering
  \caption{The mAP$_{.5}$ results of our latent object detection model on the COCO validation dataset at different timesteps.}
  \label{tab_lod}
  \vspace*{0.2cm}
  \begin{tabular}{@{}lccccc@{}}
    \toprule
    \multirow{2}{*}{model} & \multicolumn{5}{c}{timestep} \\ \cmidrule(l){2-6} 
     & 0 & 200 & 400 & 600 & 800 \\ \midrule
    YOLOv1 & 56.5 & 58.4 & 54.8 & 38.1 & 18.0 \\ \bottomrule
    \end{tabular}
\end{table}

\subsection{Language Parser}
\label{app_parser}

We utilize the noun\_chunks function in Spacy to identify the noun phrases in prompts and extract the noun within each noun phrase as the core noun. Certain noun phrases, such as time and location, do not correspond to any objects and are not relevant to our purpose. To filter out these phrases, we employ a stop-word list that includes terms such as:

top, bottom, beside, towards, front, left, right, center, middle, rear, edge, corner, periphery, interior, exterior, upstairs, downstairs, sideways, diagonal, opposite, adjacent, parallel, north, south, east, west, northeast, southeast, southwest, downward, inward, outward, lengthwise, crosswise, amidst, amongst, proximity, and vicinity.

We find that, on one hand, Spacy may make errors, while on the other hand, some noun phrases can be easily identified by the model without requiring additional guidance. Therefore, we also encourage users to annotate the noun phrases in the input prompt that require correction or fusion. This can further enhance the efficiency of the generation process and improve the quality of the resulting output.

\subsection{Algorithm}
\label{app_alg}

Here, we give the full pseudo code for our training and sampling algorithm.

\begin{algorithm}[h]
  \caption{The training process of latent object detection.}
  \begin{algorithmic}[1]
    \Require image $x_0$, bounding boxes $bbox$, class label $c$
    \While {not converge}
      \State $z_0$, prompt = Encoder($x_0$), $c$
      \State $t$, noise = random.choice($[0, T]$), random.randn(0, 1, $z_0$.shape)
      \State $z_t = \sqrt{\alpha_t} z_0 + \sqrt{ 1- \alpha_t}$ noise
      \State \_, CAM = DM($z_t$, prompt, $t$)
      \Comment{Store the CAM for latent object detection.}
      \State $bbox_p$, $conf_p$ = Detector(CAM) 
      \State $loss_{bbox}$ = $L_{bbox}(bbox_p, bbox)$
      \State $conf_{ij}$ = the center of $bbox$ is in $ij$-th cell
      \State $loss_{conf}$ = $L_{conf}(conf_p, conf)$
      \State update Detector with $loss_{bbox} + \lambda loss_{conf}$
      \Comment{Only the bbox loss and conf loss are used.}
    \EndWhile
  \end{algorithmic}
\end{algorithm}

\begin{algorithm}[h]
  \caption{Detection of Detector Guidance for Stable Diffusion.}
  \begin{algorithmic}[1]
    \Require initial noise $x_T$, text condition $c$
    \State core noun $cn$, noun phrase $np$ = Parser($c$)
    \Comment{Extract noun phrases from the text condition.}
    \For {$t = T, T-1, \cdots, T - I + 1$}
      \State $x_{t-1}$, $\text{CAM}$ = DM($x_t$, $c$)
      \Comment{Waiting for the CAM to become meaningful.}
    \EndFor 
    \For {$t = T - I, T- I-1, \cdots, 1$}
      \State $x_{t-1}$, $\text{CAM}$ = DM($x_t$, $c$, $\text{BBox}_{t+1}$, $\sigma_{t+1}$) \Comment{The correction is made in this step.}
      \State CAM = CAM[$cn$]
      \Comment{The CAMs for core nouns form a batch of input.}
      \State $\text{BBox}_t$ = Detection($\text{CAM}$)
      \Comment{Only do detection for the core nouns.}
      \State $\sigma_t$ = Otsu($\text{CAM}$)
    \EndFor \\
    \Return image = Decoder($x_0$)
  \end{algorithmic}
\end{algorithm}

\begin{algorithm}[h]
  \caption{Correction of Detector Guidance for Stable Diffusion.}
  \label{alg_dg_c}
  \begin{algorithmic}[1] %
    \Require bounding box $\text{BBox}_{t+1}$, threshold $\sigma_{t+1}$, smoothing factor $s$
    \For {block $\in$ Stable Diffusion}
      \If {block is cross-attention block}
        \State $K, Q_c, Q_{uc}, V$ = block(input)
        \State $\text{CAM}_0$, $\text{CAM}_{\text{uc}}$ = $KQ_{c}^T$, $KQ_{uc}^T$
        \State $\text{m}$ = Mask($\text{CAM}_0$, $\text{BBox}_{t+1}$, $\sigma_{t+1}$)
        \Comment{Boundary Correction}
        \State $\text{CAM}_1$ = $\text{CAM}_0 * $(1 - $\text{m}$) + $\text{CAM}_{\text{uc}} * \text{m}$
        \Comment{Conflict Elimination}
        \State $\text{CAM}_2 = \text{CAM}_1 * \frac{\text{CAM}_0.\text{max}(\text{dim}=-1)}{\text{CAM}_1.\text{max}(\text{dim}=-1)}$
        \Comment{Target Enhancement}
        \State $\text{CAM}_3$ = $\text{CAM}_2 * s$ + $\text{CAM}_0 * (1 - s)$
        \Comment{Smooth Involvement}
        \State output = $\text{Softmax}(\text{CAM}_3/\sqrt{d}) * V$
      \Else
        \State output = block(input)
      \EndIf
      \State input = output
    \EndFor
  \end{algorithmic}
\end{algorithm}

\newpage
\subsection{Multi-Related Object Benchmark}
\label{app_mro}

We use GPT4 to generate text-to-image prompts. The prompt used in GPT4 is:

``The pattern consists of visually similar yet contrasting pairs of animals or objects that can naturally coexist in a single image, emphasizing their unique attributes to create engaging and descriptive prompts. The output format should be adjective + noun and adjective + noun, such as a white cat and a black dog. Now please give me 20 prompts to meet the above requirements.''

Here, we show the whole list of the prompts in the Multi-Related Object benchmark (MRO). 

\begin{enumerate}
  \item a fluffy sheep and a bare goat
  \item a friendly koala and a watchful kangaroo
  \item a howling wolf and a purring cat
  \item a white cat and a brown dog
  \item a golden retriever and a gray wolf
  \item a regal lion and a sly fox
  \item a striped tiger and a spotted leopard
  \item a wise owl and a nimble squirrel
  \item a wild mustang and a graceful deer
  \item a robust bison and a dainty gazelle
  \item a soft bunny and a spiky porcupine
  \item a swift cheetah and a lumbering bear
  \item a cunning coyote and a timid deer
  \item a towering giraffe and a sturdy elephant
  \item a sprightly hare and a slow-moving tortoise
  \item a spotted hyena and a striped zebra
  \item a fierce falcon and a gentle dove
  \item a swift hummingbird and a perching eagle
  \item a vibrant toucan and a modest pigeon
  \item a chatty parrot and a silent owl
  \item a luminescent jellyfish and a matte sea turtle
  \item a fierce crocodile and a docile manatee
  \item a beautiful butterfly and a fluffy bee
  \item a hovering dragonfly and a perched hummingbird
  \item a wispy dandelion and a dense sunflower
  \item a red apple and a green pear
  \item a ripe peach and a tangy orange
  \item a succulent pineapple and a crisp apple
  \item a rusty robot and a delicate Muppet
  \item a futuristic drone and a traditional kite
\end{enumerate}

\newpage
\subsection{Human Evaluation}

\begin{figure}[ht]
  \centering
  \includegraphics[width=0.8\linewidth]{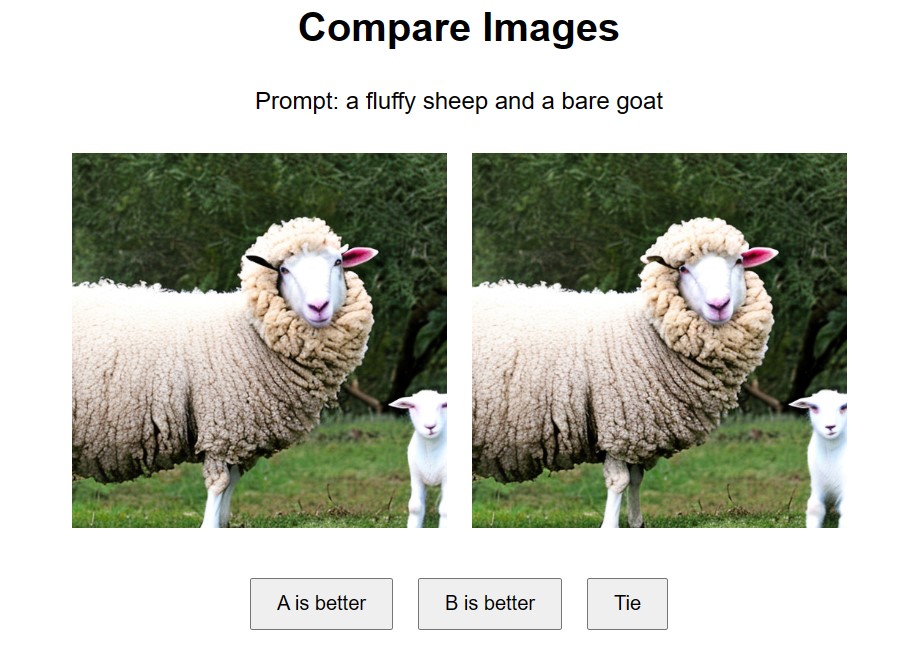}
  \caption{The website used for human evaluation.}
  \label{fig_eval_demo}
\end{figure}

In Figure \ref{fig_eval_demo}, we show the website used for human evaluation. The evaluator is required to choose one of three options - A is better, B is better, and Tie - based on the alignment between the prompt and the image.

\subsection{Project License}

Here, we present the GitHub repository addresses and the project licenses for the main open-source projects used in this paper.

\begin{table}[h]
  \centering
  \small
  \begin{tabular}{@{}lll@{}}
    \toprule
    name & GitHub & license \\ \midrule
    YOLOv1 & \url{https://github.com/yjh0410/new-YOLOv1\_PyTorch} & n/a \\
    DDPM & \url{https://github.com/hojonathanho/diffusion} & n/a \\
    DDIM & \url{https://github.com/ermongroup/ddim} & MIT license \\
    PNDM & \url{https://github.com/luping-liu/PNDM} & Apache-2.0 license \\
    dpm-solver & \url{https://github.com/LuChengTHU/dpm-solver} & MIT license \\
    SD v1.4 & \url{https://github.com/CompVis/stable-diffusion} & CreativeML Open RAIL-M \\
    SD v2.1 & \url{https://github.com/Stability-AI/stablediffusion} & CreativeML Open RAIL++-M \\ \bottomrule
    \end{tabular}
\end{table}

\end{document}